\documentclass[11pt,a4paper]{article}
\usepackage[hyperref]{acl2020}
\usepackage{times}
\usepackage{latexsym}
\usepackage{graphicx}
\usepackage{todonotes}

\usepackage{verbatim}
\usepackage{xcolor}
\usepackage{arydshln} % dashed lines in tables
\usepackage{setspace}
\usepackage{textcomp}

\usepackage{tikz}
\newcommand{\hint}[2]{
\tikzstyle{mybox} = [draw=black, fill=white, thick,
    rectangle, rounded corners, inner sep=5pt, inner ysep=10pt]
\tikzstyle{fancytitle} =[fill=white, text=black, draw=black]

\begin{center}
\begin{tikzpicture}
\node [mybox] (box){%
    \begin{minipage}{0.95\columnwidth}
{#1}
    \end{minipage}
};
\node[fancytitle, right=10pt] at (box.north west) {#2};
\end{tikzpicture}%
\end{center}
}

\newcommand{\gp}{\texttt{GePpeTto}}

\usepackage{microtype}
\usepackage{booktabs,enumitem}

\aclfinalcopy % Uncomment this line for the final submission

\title{GePpeTto Carves Italian into a Language Model}

\author{Lorenzo De Mattei \\
  University of Pisa / Pisa, Italy \\
  ItaliaNLP Lab @ ILC-CNR / Pisa, Italy \\
  CLCG, University of Groningen / The Netherlands \\
  \texttt{lorenzo.demattie@di.unipi.it} \\\And
  Michele Cafagna \\
  Aptus.AI \\ Pisa, Italy \\
  \texttt{michele@aptus.ai} \\\AND
  Felice Dell'Orletta \\
  Istituto di Linguistica Computazionale (ILC–CNR) \\
    ItaliaNLP Lab - www.italianlp.it \\
  \texttt{felice.dellorletta@ilc.cnr.it} \\\And
  Malvina Nissim \\
  CLCG, University of Groningen\\
  The Netherlands \\
  \texttt{m.nissim@rug.nl} \\\And
  Marco Guerini \\
  Fondazione Bruno Kessler  \\
   Trento, Italy \\
  \texttt{guerini@fbk.eu} \\}

\author{\bf {Lorenzo De Mattei}$^{\bullet}$$^{\diamond}$$^{\star}$$^{\dagger}$, \bf{Michele Cafagna$^{\dagger}$, Felice Dell'Orletta$^{\star}$, Malvina Nissim$^{\diamond}$, Marco Guerini$^{\ddagger}$}\\ 
~
\\
  \textsuperscript{$\bullet$}Department of Computer Science, University of Pisa, Italy \\ \textsuperscript{$\diamond$}Center for Language and Cognition Groningen, University of Groningen, The Netherlands\\
\textsuperscript{$\star$}ItaliaNLP Lab, Istituto di Linguistica Computazionale ``Antonio Zampolli", Pisa, Italy\\
\textsuperscript{$\dagger$}Aptus.AI, Pisa, Italy \\ \textsuperscript{$\ddagger$}Fondazione Bruno Kessler,  Trento, Italy\\
\normalsize{\texttt{lorenzo.demattei@di.unipi.it}}, \normalsize{\texttt{michele@aptus.ai}},\\
   \normalsize{\texttt{felice.dellorletta@ilc.cnr.it}},
   \normalsize{\texttt{m.nissim@rug.nl}}, \normalsize{\texttt{guerini@fbk.eu}}}

\date{}

\begin{document}
\maketitle
\begin{abstract}

In the last few years, pre-trained neural architectures have provided impressive improvements across several NLP tasks. Still, generative language models are available mainly for English. We develop \gp,  the first generative language model for Italian, built using the GPT-2 architecture. We provide a thorough analysis of \gp's~quality by means of both an automatic and a human-based evaluation. The automatic assessment consists in (i) calculating perplexity across different genres and (ii) a profiling analysis over \gp's writing characteristics. We find that \gp's production is a sort of \textit{bonsai} version of human production, with shorter but yet complex sentences. Human evaluation is performed over a sentence completion task,  where \gp's output is judged as natural more often than not, and much closer to the original human texts than to a simpler language model which we take as baseline.

\end{abstract}

\section{Introduction}

%We tested our approach in the evaluation of source-specific headline generation. 

Language Models (LMs) based on pre-trained architectures such as BERT \cite{bert} and GPT-2 \cite{radford2019language} have provided impressive improvements
across several NLP tasks. While for BERT-based architectures several monolingual models %for languages 
other than English have been developed, language-specific implementations of generative pre-trained
transformer based models, such as GPT-2, are not widely available yet. As a contribution to fill this gap, we developed \gp, the first generative language model for Italian, using the original GPT-2 as a blueprint. 

%\textcolor{blue}{After presenting the data, architecture and configuration used to build \gp,} 
The evaluation of generated text is known to be intrinsically difficult \cite{gatt2018survey}; we adopt here an encompassing approach, performing both automatic and human-based evaluations. The automatic assessment consists in two strategies: the first involves calculating perplexity across different language models trained on various datasets representing different genres. This serves to understand %should tell us 
how good \gp~is as a language model, and how much it captures %resembles 
the various genres. The second one is a profiling analysis where, %. In other words, 
%through 
by means of a series of linguistic features, we capture some of \gp's writing characteristics, and compare them to those of the %original 
data %\gp~
it was trained on. Finally, the human evaluation %consists in human judgements 
is performed 
over a sentence completion task  where \gp~is evaluated against gold standard sentences as well as a simple Markov-based baseline.

We make the model available to the community: \url{https://github.com/LoreDema/GePpeTto}.

\section{\gp}
\label{sec:GP}

\gp~was trained using the original settings of GPT-2 %put together 
on a collection of Italian texts %for 
amounting to %a total of 
almost 13GB. Details on data and  model's parameters are provided in the following sections.

\subsection{Data}

%We put together a 
The training set comprises two main %different 
sources. The first one is a dump of Italian Wikipedia (November~2019), consisting of 2.8GB  of text. The content was extracted using the Wikiextractor tool \cite{wikiextractor}. The second one is the ItWac corpus \cite{itwac}, which amounts to 11GB of web texts. This collection provides a mix of standard and less standard Italian, on %as well as 
a rather wide chronological span, with older texts than the Wikipedia dump (the latter stretches only to the late 2000s).

Minimal processing was applied to the texts. All Wikipedia documents %texts 
were prefixed by the token ``Wikipedia" followed by the page's title words. All ItWac texts were introduced by the token ``Links" followed by the webpage address the text was coming from. For all texts in both collections, end of document was marked with the string \verb+<|endoftext|>+, as done for the original GPT-2 training set \cite{radford2019language}.
%Wikipedia: Wikipedia Titolo

%Itwac: Link URL

%Marcata fine testo con special token.

%<|endoftext|>

%\todo{non diciamo niente di test sets per i due dataset qui, ma neanche nella section perplexity. da qualche parte deve andare}

\subsection{Model}

\gp~was trained on 4 GPUs Tesla T4, using the PyTorch implementation available on huggingface.\footnote{\url{https://huggingface.co}.} The model's size corresponds to GPT-2 small, with 12 layers and 117M parameters \cite{radford2019language}. The vocabulary size is 30,000, the batch size is 32 and the block size is 100. We use 10,000 warmup steps, the adam optimiser, and an initial learning rate of 5e-5. We train the model for 620,000 steps.

%hardware: 4 gpu tesla 4
%implementation: hugging face pytorch
%model size: gpt-2 small 117M, 12 layer
%vocab size: 30k
%batch size: 32
%block size: 100
%warmup steps: 10000
%optimizer: adam
%initial learning rate: 5e-5 
%steps: 620k

\subsection{Examples}

Through some selected examples, we illustrate here \gp's generation behaviour. We prompt it with different inputs, and show how the generated text is indeed style-compliant with the prompts. In the first example, we start the prompt with the `keyword' ``Wikipedia", priming \gp~to reproduce text in the style and format of Wikipedia pages (recall that the tag ``Wikipedia" was preposed to all Wikipedia training data).

% \textcolor{blue}{Some examples, generated over the two tags ``Wikipedia”  and ``Links” used for training, are provided in Table XX. The conditioning text is highlighted in red.} %\textbf{\textcolor{red}{@LORE O QUALCUNO: few (cherry-picked) examples of produced text. Conditioning on links (if we have two very stile-specific links let's use them) and wikipedia. For Wikipedia, let's try to use a fictional state or somenthing similar, e.g. "$<wiki>$ Il Burmini,  ufficialmente Repubblica Democratica del Burmini, è uno stato situato nel sud" }}

%\textcolor{red}{visto che links è troppo generico e non triggera uno stile specifico oltre a wikipedia ho messo due prompt che triggerassero stili specifici (forum, leggi)}

% \hint{\bigskip ``Wikipedia La Repubblica dell'Isola di Catan \textbackslash n\textbackslash nLa Repubblica dell'Isola di Catan \`{e} la pi\`{u} grande area protetta del Per\`{u}, anche se \`{e} il posto pi\`{u} popolato del paese.\textbackslash n\textbackslash nLa capitale Catan \`{e} la città pi\`{u} grande del paese e il principale centro turistico del paese. La citt\`{a}, essendo cos\`{\i} capoluogo del dipartimento, si estende per 1.189 km$^{2}$. 
% %Con un'estensione di 1.228 km$^{2}$, conta circa 1.686 km$^{2}$, la maggior parte della superficie di Pana si trova nella Provincia del paese.
% "}{\parbox{.9\columnwidth}{prompt: ``Wikipedia La Repubblica dell'Isola di Catan"}}

% generated: Wikipedia Burminidilus falcheni

\smallskip

\hint{\bigskip ``Wikipedia La Repubblica del Burmini \textbackslash n\textbackslash nLa Repubblica del Burmini (in francese ``La Repubblica del Burginais"), originariamente repubblica di Francia, venne creata nel 1859 insieme al Regno d'Italia, con capitale Bourges, per promuovere l'unificazione dei due territori della Francia. Le città della Francia si unirono nel 1877 ed i cinque stati facenti parte dell'impero erano Francia, Italia e Savoia."
}{\parbox{.8\columnwidth}{prompt: ``Wikipedia La Repubblica del Burmini"}}

\noindent In the second example, we prompt \gp~ with the same text ``\textit{La Repubblica del Burmini}" but preposing to it the `keyword' ``Links" followed by the web address of an Italian newspaper, priming \gp~to reproduce a text in the style of a news article. Interestingly the output is in news style but the prompt is interpreted as the header of the news, i.e., `place and date'.
 
\hint{\bigskip ``Links http://www.corriere.it La Repubblica del Burmini  - 14 gennaio 2003. Nota: Per gli ambiti regionali, le Regioni dovranno determinare in un solo anno le modalità di trasporto, tra gli altri, degli studenti del secondo ciclo, e le modalità per pagare eventuali supplementi a carico del servizio sanitario nazionale. Si tratta, in altre parole, di un decreto legislativo con un semplice riferimento alla Regione Lombardia.}{\parbox{.8\columnwidth}{prompt: ``Links http://www.corriere.it La Repubblica del Burmini"}}

% I bruminidi, detti anche jocavidi ("Flogsburg night walts" in lingua inglese, ) sono pesci di acqua dolce dell'Africa orientale.

\noindent In the following three examples we can see that even without using any specific keywords, \gp's is easily style-primed by clear prompts. In the first case, we use as input a string prompting forum-like language, in the second one, we use legal terms in the prompt, and in the last one, we prompt \gp~ with more sophisticated language, which the model appears to readily pick up in its generated text.

\hint{\medskip ``Piero DM wrote: $>$Una domanda sul lavoro a tempo pieno dovrebbe far sorridere......$<<$ Scusate, ma vorrei avere una vostra opinione."}{prompt: ``Piero DM wrote:"}

\hint{\medskip ``Visto l'articolo 4 del dpcm 29 dicembre 1999, con il quale viene disposto l'inquadramento tra il personale interessato alle operazioni e le RSU per l'applicazione del quarto comma del predetto articolo 4, paragrafo 1, lettera a) del presente accordo, nei ruoli dei giornalisti, delle altre amministrazioni dello Stato, delle agenzie ed aziende autonome; Vista la legge 29 marzo 1982, n. 93, e successive modificazioni,}
%con la quale è stato approvato il testo unico delle disposizioni legislative vigenti in materia di tutela e sicurezza del lavoro, di cui al decreto legislativo del Capo provvisorio dello Stato 14 novembre 1993, n. 670/X;"}
{prompt: ``Visto l'articolo 4 del dpcm:"}

\hint{\medskip ``l'intricato esistenzialismo  diventa il motore di quella riflessione metafisico-filosofica di cui il primo pensatore si avvale. L'altruismo di fondo appartiene quasi completamente alla filosofia; l'acuto verso la razionalità senza limite.}{prompt: ``l'intricato esistenzialismo"}

\section{Automatic Evaluation}

\gp~is trained as a language model for Italian. To assess its closeness to actual Italian texts, we calculate perplexity on a variety of sources, including a small leave out test set (1\%) of \gp~s  training corpus (Section~\ref{sec:perplexity}). In addition, we explore more in depth \gp's linguistic profile by comparing its production with human-written texts along a series of linguistic features (Section~\ref{sec:profiling}).

\subsection{Perplexity}
\label{sec:perplexity}

As a first evaluation, we are interested in understanding the quality of \gp~as a language model in its own training domain. As a second evaluation we want test its performance at zero-shot domain transfer (i.e. language modeling of a different domain). We use perplexity as a measure of language modelling performance. The different domains we consider, and the relative corpora we use, are as follows:

\begin{itemize}

\item own domains: Wikipedia and ItWac;

\itemsep 0pt

\item legal domain: a corpus of Italian laws scraped from EUR-Lex\footnote{https://eur-lex.europa.eu/} (tables excluded);

\item news: a corpus of articles from the online versions of two  major Italian newspapers, namely \textit{la Repubblica}\footnote{\url{https://www.repubblica.it}} and \textit{Il Giornale}\footnote{\url{https://www.ilgiornale.it/}
}  \cite{invisible-dema-2020};

\item social media: a corpus of forum comments  \cite{maslennikova2019quanti}.

\end{itemize}

\noindent Perplexity scores are reported in Table~\ref{tab:perplexity}. As we could expect, \gp~%tends to 
performs better on its own domains, with Wikipedia being the best of the two. %The lower performance on ItWac compared to Wikipedia, 
Although ItWac is four times bigger than Wikipedia, the lower performance on the former might be due to the fact this corpus is open domain with a large diversity of styles, while Wikipedia is more `standardised'. Consistently with this hypothesis, we observe a similar trend in `out-of-domain' testing, where \gp~ performs better on domains with a well coded style, namely legal documents. On domains with less coded styles, such as news and especially forum comments,  we observe a drop in performance. 

If we compare perplexity scores with the original English GPT-2 small model, we see that \gp's results are slightly worse on the own domain corpora, which could be due to the smaller size of the training set. Out-of-domain perplexity scores are comparable between the two models. 

%(ii) the lower quality of training data, (iii) \gp~might be under fitted as we do not have enough computational power to further train it in a reasonable amount of time.
%Nevertheless in general from these results we can expect \gp to perform well on a series of Italian domains, except maybe for sparse domains such as forum comments or Social Media. 
%For this kind of data, \textcolor{blue}{coming from Social Media platforms}, fine-tuning may help.\todo{io forse questo commento lo toglierei}

\begin{table}[tbh]
    \centering
    \begin{tabular}{l|r}
    \toprule
    \textsc{domain} & \textsc{perplexity}\\ 
    \midrule
    Wikipedia & 26.1052 \\ 
    ItWac & 30.3965 \\
    \midrule
    Legal  & 37.2197 \\
    News & 45.3859 \\
    Social Media & 84.6408  \\
    \bottomrule
    \end{tabular}
    \caption{Perplexity of \gp~over several in-domain and out-of-domain corpora.}
    \label{tab:perplexity}
\end{table}

\subsection{Linguistic Profiling}
\label{sec:profiling}

For our second evaluation, we used Profiling-UD \cite{profilingud-brunato-2020}, a tool for the automatic analysis of texts that extracts several linguistic features of varying complexity.
%The linguistic features are automatically extracted using Profiling-UD \cite{profilingud-brunato-2020} and they range
These features range from raw text properties, such as  average length of words and sentences, to lexical, morpho-syntactic, and syntactic properties, such as part-of-speech (POS) distribution and inflectional properties of verbs. 
% depth of subordination. %, which were extracted from different levels of linguistic annotation.
%These features capture several linguistic phenomena ranging from the average length of words and sentence, to morpho--syntactic information both at the level of POS distribution and about the inflectional properties of verbs. 
More complex aspects of sentence structure are derived from syntactic annotation, and model global and local properties of parsed tree structure, such as the order of subjects/objects with respect to the verb, the distribution of syntactic relations, and the use of subordination.

\begin{table}
\centering
\begin{tabular}{l|rr|rr}
\toprule
& \multicolumn{2}{c}{Original} & \multicolumn{2}{c}{\gp} \\ \midrule
Feature & $\mu$ & std & $\mu$ & std \\ \midrule
CPT & 4.809 & 0.959 & 4.750 & 1.127\\
TPS & 32.302 & 28.322 & 20.382 & 11.127 \\
TPC & 12.393 & 11.504 & 10.711 & 8.529 \\
LL$_{max}$ & 13.290 & 13.370 & 8.922 & 6.112 \\
LL$_{avg}$ & 2.555 & 1.002 & 2.373 & 0.676 \\
\bottomrule
\end{tabular}
\caption{\label{tab:profiling_feat}Main linguistic features considered in our analysis. CPT = chars per token, TPS = token per sentence, TPC = tokens per clause, LL = links length.} %All differences are statistically significant with $p<0.001$.
\end{table}

In our analysis we focus on two macro aspects of \gp's output, namely lexical complexity and syntactic complexity, and compare them to human productions. To do so, we rely on a selection of Profiling-UD's features which we use as proxies for the macro-aspects that we consider.

We run the profiling analysis on a sample of both gold and generated texts. For gold, we randomly sample the test set for a total of about 19k sentences. For \gp, we picked  the first token from each of the 19k gold sentences, and used it as a prompt to the model. These are the generated texts that we profile.

\paragraph{Lexical complexity.}

We proxy lexical complexity with the number of characters per word, overall frequency of tokens, also with reference to an external dictionary, and POS distribution. 

The number of characters per token (CPT), which indicates whether shorter (usually more common) or longer (usually more complex/specialised) words are used, is completely comparable across the original (4.80, std=0.96) and \gp's (4.75, std=1.13) language models -- see Table \ref{tab:profiling_feat}. This suggests that the complexity of the used vocabulary is not that different.

We compute a reference dictionary of token frequency on ItWac ($\approx$1.5~billion tokens), and compare observed token frequency in both gold and generated text to this reference. We observe that in gold sentences, each token has a probability of 0.912 to be in the top 5\textperthousand~of most frequent tokens. In the generated sentences, the probability grows to 0.935, suggesting that \gp~is more likely to use more frequent words rather than rarer ones. 
%This is also confirmed if we look at what proportion of words is used which has a reference frequency $<10,000$: 1.3\% for \gp~and 1.6\% 
% it seems that the proportion of infrequent words (as calculated on an external reference corpus) is much higher in gp than in the original data. Weird.
This observation is in line with previous research which showed that for Nucleus Sampled texts, such as those produced by GPT-2, all tokens come from the top-p\%, since the long tail is cut off, while for human produced texts, the probability of all tokens being drawn from the top-p\% of the language distribution goes to zero as document length increases \cite{gehrmann2019gltr,zellers2019defending}.

Regarding POS distribution, we observe that while for most POS tags usage is comparable, for a few others the two language models differ.
%\todo{add footnote on automatic tagging can have mistakes?}. 
The latter are, specifically, auxiliaries and proper nouns, which \gp~tends to overgenerate in comparison to the original model, and adjectives, which \gp~instead uses less than in the original texts. This is observable also for nouns and verbs, but the differences are relatively minimal. Conjunctions are also overall less frequent in \gp. See Table \ref{tab:profiling_pos} for  details.

\begin{table}
\centering
\begin{tabular}{l|cc|cc}
\toprule
& \multicolumn{2}{c}{Original} & \multicolumn{2}{c}{\gp} \\ \midrule
POS & $\mu$ & std & $\mu$ & std \\ \midrule
%	&  \\  %Orig_{$\mu$} &	Std\_Orig &	Mean\_\gp &	Std\_\gp
AUX	& 0.032	& 0.041 &	0.040 &	0.051 \\
PROPN & 	0.070 &	0.105 &	0.081 &	0.125 \\
PUNCT & 	0.148 &	0.103 &	0.153 &	0.105 \\
DET	 & 0.140 &	0.071 &	0.143 &	0.078 \\
NUM	 & 0.031 &	0.072 &	0.032 &	0.064 \\
ADP	 & 0.139 &	0.070 &	0.138 &	0.077 \\
PRON & 	0.037 &	0.053 &	0.036 &	0.058 \\
SCONJ & 	0.008 &	0.020 &	0.008 &	0.023 \\
NOUN & 	0.179 &	0.082 &	0.172 &	0.087 \\
VERB & 	0.079 &	0.059 &	0.075 &	0.065 \\
ADV	& 0.042	& 0.060 &	0.039 &	0.063 \\
CCONJ &	0.027 &	0.034 &	0.024 &	0.037 \\
ADJ	& 0.063	& 0.058 &	0.055 &	0.062 \\
\bottomrule
\end{tabular}
\caption{\label{tab:profiling_pos}POS considered in our analysis.} %\textcolor{red}{All differences are statistically significant with $p<0.001$}} 
\end{table}

\paragraph{Syntactic complexity.}

At the level of syntax, we proxy complexity by the number of tokens per sentence, and the number of tokens per clause. %\textcolor{red}{calculated in terms of the average number of tokens per clause, where the number of clauses corresponds to the ratio between the number of tokens in a sentence and the number of either verbal or copular heads}. 
We also look at the length of a dependency link, that is calculated as the number of words occurring linearly between the syntactic head and its dependent (excluding punctuation dependencies). The value associated with this feature corresponds to the average value extracted for all dependencies in a text. 
This information is complemented with  
the feature \textit{Maximum dependency link} corresponding to the longest dependency link for each sentence.

When comparing the number of tokens per sentence (TPS, Table~\ref{tab:profiling_feat}), we see that it's much lower for \gp's production rather than for human texts (20.4 tokens per sentence on average for \gp~vs 32.3 for gold texts),%, for example), 
indicating that \gp~generates shorter sentences. Contextually, we also observe that \gp's generated sentences exhibit less variation in length (smaller STD) than human sentences (larger STD). 

The difference in number of tokens at the clause level is relatively smaller, with clauses of length 12.4 in human texts vs 10.7 in \gp~(TPC, see Table~\ref{tab:profiling_feat}). Considering that a clause is proxied by the presence of a verbal/copular head, it seems that sentences produced by \gp, though shorter, are similar in complexity given the proportional distribution of verbal heads.

The above values taken together might suggest that while complexity at the macro level (sentence length) is higher for natural sentences, at the micro level  (clause length) %of smaller chunks, 
complexity of \gp's generations and human texts is more similar. While this intuition will require further linguistic analysis, it seems to be confirmed by the data we have if we look at the length of syntactic links. This feature proxies quite well syntactic complexity, since it indicates how maximally far (and how far on average) a dependent and its head are within a sentence. Both the maximum length and the average length are higher for human texts (LL$_{max}$ and LL$_{avg}$, see Table~\ref{tab:profiling_feat}). However, if we look at them proportionally to sentence length, we find that they are absolutely comparable: normalising the longest link by the number of tokens per sentence (LL$_{max}$/TPS), we obtain basically the same value for gold (0.411) and for \gp~(0.438). This suggests that \gp~produces somewhat shorter sentences, but their internal complexity relatively corresponds to the internal complexity of the longer sentences produced by humans.

\section{Human evaluation}

We also test \gp's ability  to  generate  Italian texts through a sentence completion task. The automatically generated sentences are presented to human subjects for evaluation on perceived naturalness and compared to gold ones and to a baseline.

While  the original (gold) texts represent an upperbound for \gp, we do not actually have a lowerbound against which the quality of \gp~can be assessed. To provide a comparison, we train a simple Markov model that would be able to generate text  and use it as our baseline. Since the size of a Markov model dramatically grows  with its vocabulary size, we use 1 million randomly sampled sentences from the same training-set used for \gp. We train a Markov chain generator using the \texttt{markovify}\footnote{\url{https://github.com/jsvine/markovify}.} implementation with state size 2, then we generate synthetic texts starting from the last 2 tokens of same prompts used for \gp.

\subsection{Tasks}

Human subjects are asked to perform two evaluation tasks. One is a comparative ranking task, where subjects are asked to rank three portions of text (produced by gold, \gp, baseline) according to perceived naturalness. The other is a classification task, where subjects are asked to tell, according to their intuition, if a portion of text, seen in isolation, is automatically generated (\textit{yes, no, can't tell}).

%three types of sentences according to length: short ($<10$), medium ($>10 >20$), long ($>20$). For each set, cutting at three stages: one third, half, three thirds.

\paragraph{Experimental design.}  The experiment includes 12 conditions of the stimulus material in a 4x3 design. One level (A) with three conditions is given by \{gold,\gp, baseline\}. The second level (B) is the \texttt{prompt+completion} combination that results in 4 conditions \{5+5, 5+10, 10+5, 10+10\}. We use 100 different prompts (randomly selected gold sentences truncated at 5 and 10 tokens). Each of the 100 prompts enters each of the 12 conditions of the 4x3 design, for a total of 12 different stimuli. Basically, each 5 or 10 tokens prompt is  completed with 5 or 10 tokens coming either from  gold, \gp, or the baseline model. Table~\ref{tab:examples} shows an example of all the stimuli deriving from the same 5- or 10-token prompt.

%We have opted for an between subject evaluation set up.
Each subject is assigned either to the ranking or to the classification task. 

In ranking, we opt for a between subject evaluation set up by assigning each subject to one of the (B) conditions and offer the three versions of (A) to be ranked. For example, one subject is asked to evaluate all the 100 prompts in the 5+5 configuration (dimension~B) for the three realisations, i.e., gold, \gp, and baseline (dimension~A). 

% sostanzialmente creare 12 sets e each subject is assigned to one set only. In ogni set ci va un solo completamento di un prompt (e.g. solo 5+5 markov).

% I 12 completamenti di un single prompt vanno smistati nei 12 sets.

For the classification experiments, we again opt for a between subject evaluation set up, this time by assigning each subject to one of the 12 conditions, randomly picked up for each prompt. In other words, we make sure that each subject is exposed to only one completion per prompt, randomising prompt order. By seeing only one (out of 12) realisation per prompt, each subject sees a given prompt only once and we can therefore avoid cross-comparison effects of different completions of the same prompt, which could otherwise potentially lead again to an implicit ranking task.

%Split generated sentence and its original into different sets for different annotators.

%In the single evaluation procedure, the three different types of sentences (geppetto, markov, gold)

%One original sentence (before chopping) has a total of 12 versions. 

\begin{table*}[!t]
    \centering
    \begin{tabular}{lp{14cm}}\toprule
       \multicolumn{2}{l}{\hspace*{2.5cm}5 token prompt: Mentre per quanto riguarda gli} \\
        \multicolumn{2}{l}{\hspace*{2.33cm}10 token prompt: Mentre per quanto riguarda gli accordi per la fornitura di}\\

         \midrule
    \multicolumn{2}{c}{Gold}\\
    \midrule
    
    5+5 & Mentre per quanto riguarda gli accordi per la fornitura di\\
    5+10 & Mentre per quanto riguarda gli accordi per la fornitura di latte, in scadenza questa\\
    10+5 & Mentre per quanto riguarda gli accordi per la fornitura di latte, in scadenza questa\\
    10+10  & Mentre per quanto riguarda gli accordi per la fornitura di latte, in scadenza questa settimana, Alemanno ha detto\\

        \midrule
    \multicolumn{2}{c}{\gp}\\
    \midrule
     
    5+5 & Mentre per quanto riguarda gli emendamenti, fa presente che il\\% \cdashline{1-2}
    5+10 & Mentre per quanto riguarda gli emendamenti, fa presente che il suo gruppo non ha sottoscritto\\ %\cdashline{1-2}
    10+5 & Mentre per quanto riguarda gli accordi per la fornitura di beni e servizi, i fatti\\
    10+10  & Mentre per quanto riguarda gli accordi per la fornitura di beni e servizi, i fatti in suo possesso hanno come\\

    \midrule
    \multicolumn{2}{c}{Markov-based baseline}\\
    \midrule
    
        5+5 & Mentre per quanto riguarda gli aspetti più significativi del mondo\\
    5+10 & Mentre per quanto riguarda gli aspetti più significativi del mondo editoriali, con priorità di sviluppo\\
    10+5 & Mentre per quanto riguarda gli accordi per la fornitura di biciclette elettriche a 48 bit\\ %\cdashline{1-2}
    10+10  & Mentre per quanto riguarda gli accordi per la fornitura di biciclette elettriche a 48 bit (281,5 trilioni di operazioni e\\
\bottomrule
    \end{tabular}
    \caption{\label{tab:examples}Example outputs (stimuli) for different prompt lengths of the same original sentence.}
\end{table*}

\paragraph{Material.} The materials are prepared as follows: we have selected 100 random documents/sentences and have cut them at their 5 first tokens and also their 10 first tokens. Each 5-token and 10-token prompt was given to \gp~ and baseline so that the models could continue the text. 

For each prompt, we obtain one single generated text by the two automatic models and chop them at 5 or at 10 tokens. In other words, each chopped version is derived from the same generated output which is just cut at different lengths. 

We cut the sentences (including the original one) to control for the effect of text length. Indeed, we observed in Section \ref{sec:profiling} that \gp~generates shorter sentences than humans, which could represent a strong bias in evaluation. In Table~\ref{tab:examples}, we show examples of all the possible stimulus material configurations according to the \texttt{prompt+completion} conditions of level~(B).

\begin{figure}[!p]
\begin{center}
\begin{minipage}{\columnwidth}
% \centering
\begin{center}
\fbox{\parbox{.905\columnwidth}{{\includegraphics[width=.9\columnwidth]{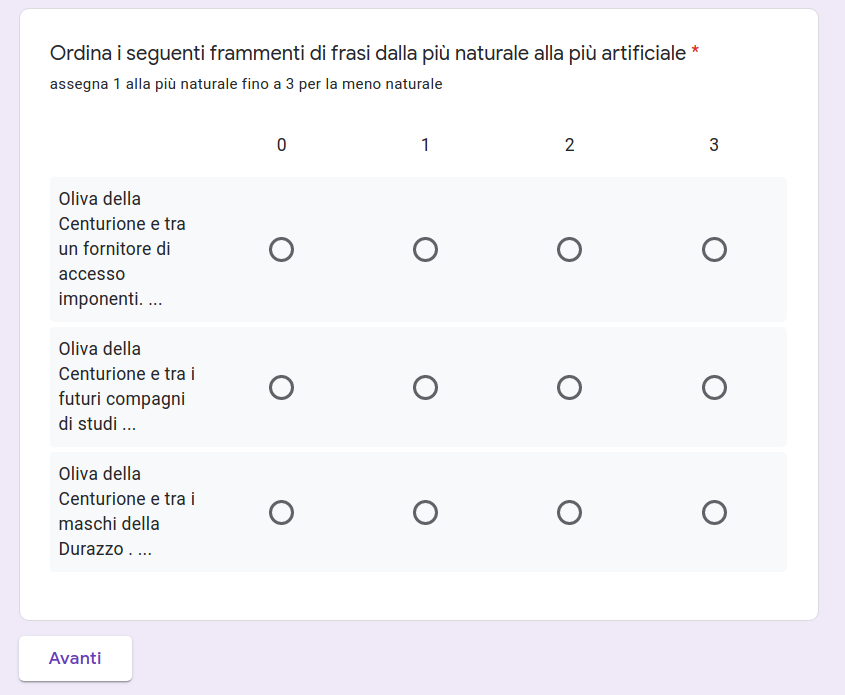}
%\end{minipage}%

\medskip

%\begin{minipage}{.9\columnwidth}
%  \centering

\includegraphics[width=.9\columnwidth]{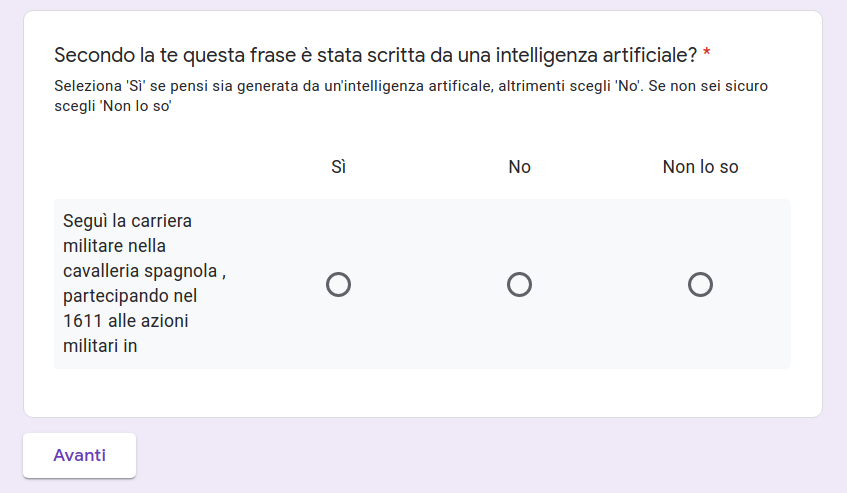}}}}
\end{center}
\end{minipage}
 \caption{Annotation interfaces for the ranking and classification tasks. \label{fig:interface}
}
 \end{center}
\end{figure}

%We cut the sentences (including original) to control for the effect of text length on evaluation. Indeed, we have observed that automatic models different models can generate different output lengths \textcolor{blue}{as seen in Section \ref{sec:profiling}} and the longer the sentence the higher the probability to be detected as natural or artificially generated). In Table~\ref{tab:examples} we show a few examples.\todo{not sure about this whole para wrt previous one}

\paragraph{Instructions and subjects.} For both the ranking and classification experiments, subjects were told that they will have to evaluate excerpts of text along a `more natural vs. more artificial' dimension. All stimuli used in both scenarios are the same.

For the ranking scenario, subjects were asked to ``\textit{rank the given examples from the most natural to the most artificial"}, where the inputs are three texts (gold, \gp, baseline), all starting with the same prompt, thus the same five or ten tokens.

For the classification scenario, subjects saw instead the portions of text in isolation, and could answer \textit{yes}, \textit{no}, or \textit{can't tell} to the question ``\textit{according to your intuition is this sentence written by an artificial intelligence?}''. 

%The tasks were prepared and carried out using Google Forms; in Figure~\ref{fig:interface} we report a sample snapshot of the interface for both task.

A total of 24 unique subjects (12 females) carried out the tasks using Google Forms (see Figure~\ref{fig:interface} for a snapshot of the interfaces.) Twelve subjects (6 females) were assigned to Task~1 and the others to Task~2. Each subject evaluated 100 cases, and each case was evaluated by three different subjects. %A total of 200 cases was assessed in each task. \todo{Perchè 200?}

%\textcolor{red}{\paragraph{Subjects.} We then recruited X subject through the Y platform. THe average age was, x where male and x female, the education level was xx or above. }

%\item fine-tuning: headline generation with human evaluation, rank out of three (gold, \gp, pointer)

%\subsection{Baseline}

%While our upperbound is constituted by the original (gold) sentences which we can use in evaluation, we do not actually have a lowerbound against which the quality of \gp~can be assessed.
%To provide a comparison, we trained a simple Markov model that would be able to generate text. Since the size of a Markov model dramatically grows  with its vocabulary size, we used 1 million randomly sampled sentences from the same training-set used for \gp. We trained a Markov chain generator using the \texttt{markovify}\footnote{\url{https://github.com/jsvine/markovify}.} implementation with state size 2, then we generated synthetic sentences starting from the last 2 tokens of same prompt used for \gp.

\subsection{Results}

First, we discuss the results of our human evaluation separately, with observations related to the ranking task and observations related to the classification task. Subsequently, we knit together the two outcomes to draw a wider picture of how humans assess the quality of \gp's output.

\paragraph{Ranking}

\begin{figure}[ht!]
\centering
\includegraphics[width=1\linewidth]{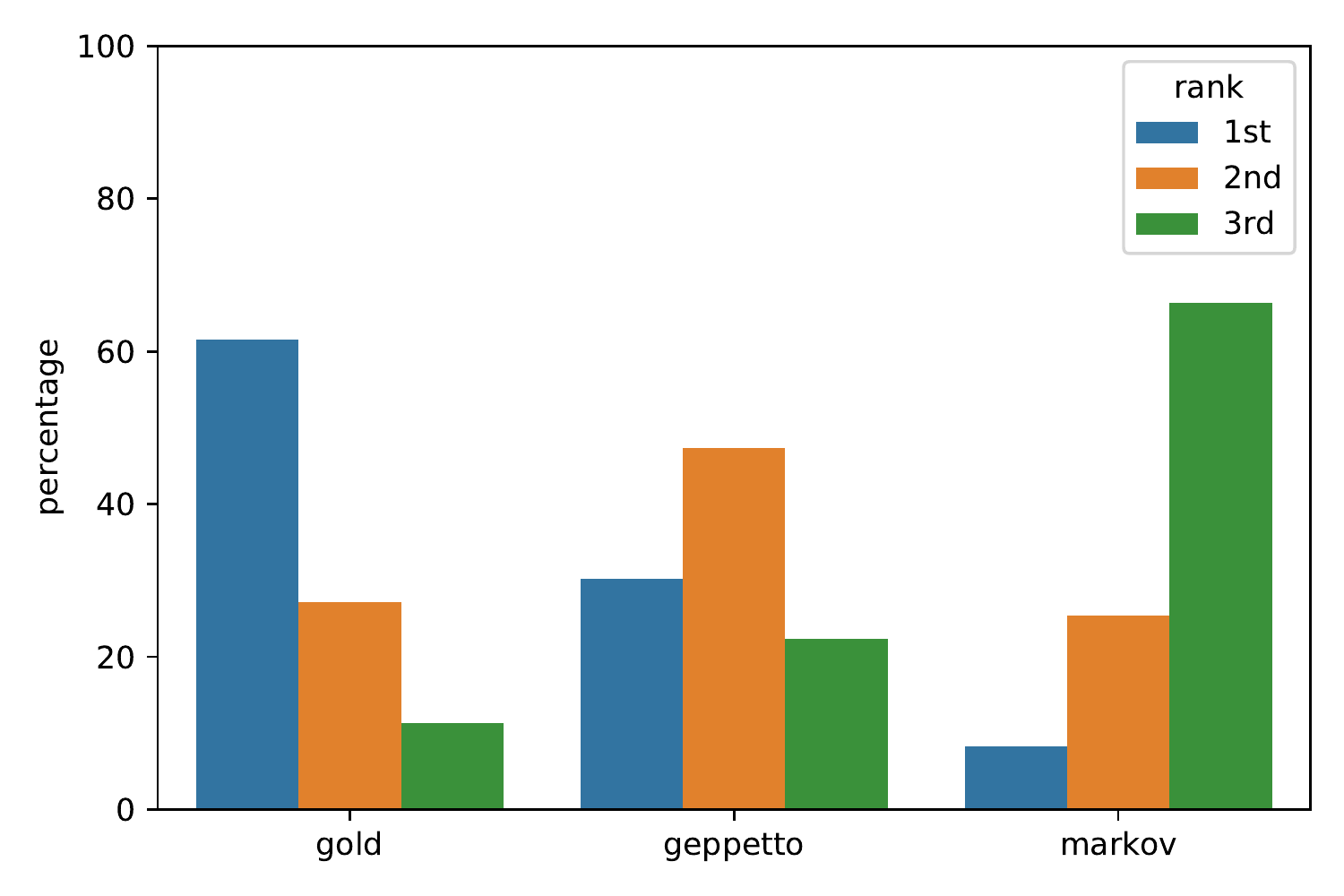}
 \caption{\label{fig:rank}Ranking results for the three models}
\end{figure}

Overall, results show that the most frequently chosen completion is the gold one, followed by \gp~and then the Markov baseline, but the baseline is far more distant from \gp~than \gp~from gold (Figure~\ref{fig:rank}). 
If we look at results in more detail (see Table~\ref{tab:rank}), based on the variable that we have considered in the experimental set up, namely length of input and continuation as well as overall sentence length, we observe that the order of preference for gold is 10+10, then 5+10, then 10+5, and lastly 5+5, while for the automatic models the order is 5+5, 10+5, 5+10, and then 10+10, suggesting the following.
%while for the generation models, the order of preference is 5+5, then 10+5, then 5+10, then 10+10. 

\begin{table*}[!t]
\centering
\begin{tabular}{l|ccc|ccc|ccc|ccc} \toprule
model & \multicolumn{3}{|c|}{5+5} &\multicolumn{3}{|c|}{5+10} &\multicolumn{3}{|c|}{10+5} & \multicolumn{3}{|c}{10+10}\\
\midrule
     & 1$^{st}$ & 2$^{nd}$ & 3$^{rd}$ & 1$^{st}$ & 2$^{nd}$ & 3$^{rd}$ & 1$^{st}$ & 2$^{nd}$ & 3$^{rd}$ & 1$^{st}$ & 2$^{nd}$ & 3$^{rd}$ \\ \midrule

  Gold  & 54 & 30 & 16 & 62 & 31 & 7 & 60 & 27 & 13 & 70 & 21 & 9  \\ \midrule
  \gp   & 34 & 43 & 23 &  30 & 46 & 24 & 33 & 43 & 24 & 23 & 59 & 18  \\ \midrule
  Markov & 12 & 27 & 61 &  8 & 23 &  69 & 7 & 30 & 63 & 7 & 20 & 73  \\
\bottomrule
\end{tabular}
\caption{\label{tab:rank} Percentages of ranking results according to the various stimulus material conditions.}
\end{table*}

First, the shortest the sentence, the hardest it is to discriminate between gold and generated text; indeed, the 5+5 condition is the one that results best for the two models and worst for gold. 

Second, when the sentence is the longest (10+10), it is easiest for the subjects to discriminate the gold from the generated sentences. It is also interesting to note that in this condition we observe the largest gap between the two generation models, with \gp~getting ranked higher than Markov more than in the other conditions.

Third, at equal sentence length (15 tokens) the situation is a bit more fuzzy, but we can observe a slight tendency where it is easier to spot as automatically generated the 5+10 rather than 10+5 cases. This, in combination with the previous observation, seems to imply that the longer the generated text, the easier it is to figure out which texts are automatically produced, which makes sense, since there is more `space' for the models to make mistakes.

% \begin{table}\todo{regarding ranking patterns: decide if to keep it or not. in case, needs a couple of comments.}
% \centering
% \begin{tabular}{c|rr} \toprule
% rank & freq &  \% \\ \midrule
% $O>G>M$     & 465	&    38.75 \\
% $G>O>M$  & 264	  &  22.00\\
% $O>M>G$      & 194	 &   16.17 \\
% $G>M>O$  & 140	 &   11.67\\
% $M>O>G$  & 88	 &   7.33\\
% $M>G>O$  & 49	 &   4.08\\
% \bottomrule
% \end{tabular}
% \caption{Ranking patterns\label{tab:patterns}}
% \end{table}

% For gpt-2 and Markov: 1, 3, 2, 4. 
%When the completion is short, the models get ranked higher, independently of the input's length. When the completion is longer, it is easier 

\paragraph{Classification}

\begin{figure}[t!]
\centering
\includegraphics[width=1\linewidth]{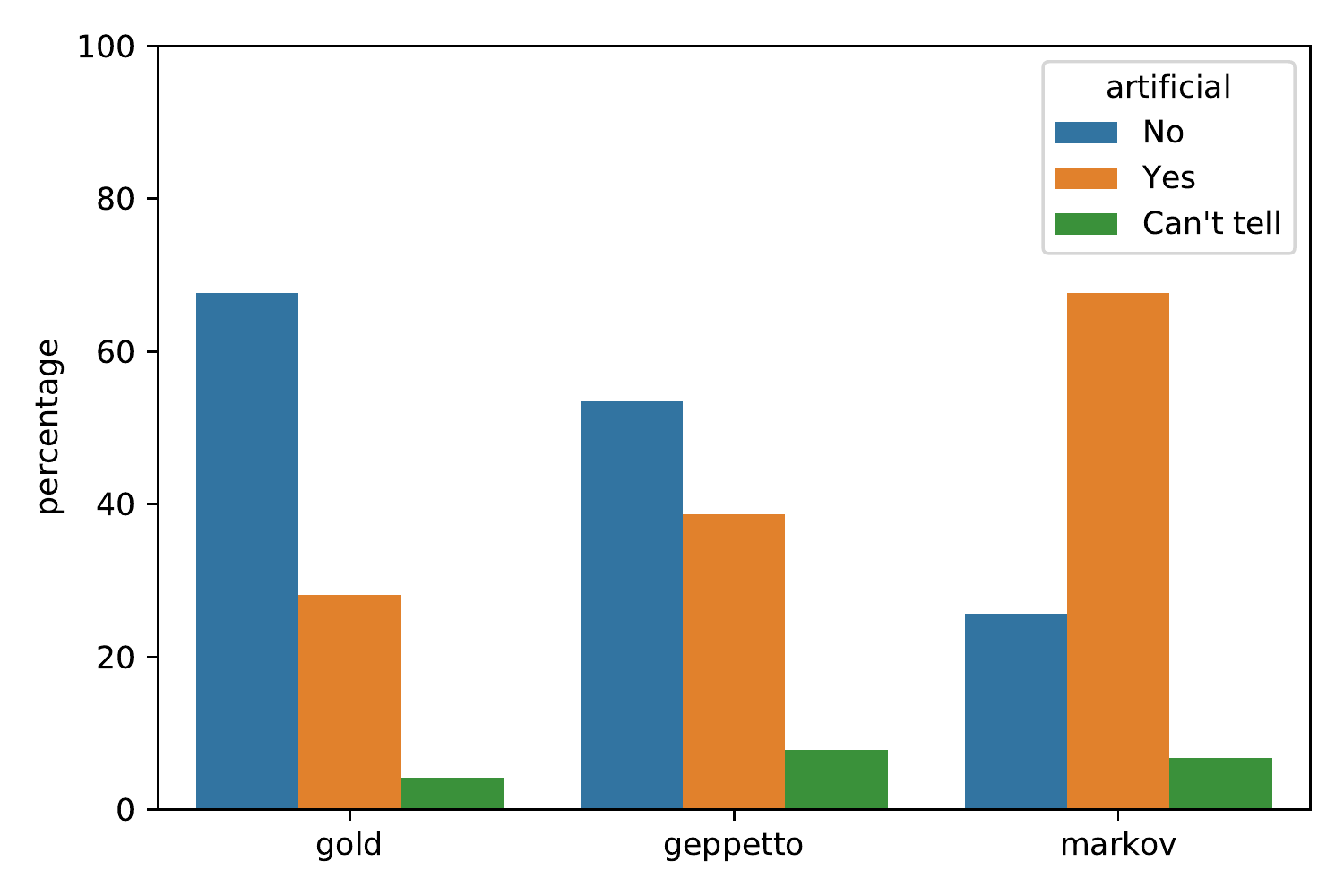}
 \caption{\label{fig:class}Classification results for the three models}
\end{figure}

Overall, results show that across all conditions, gold sentences are most often rightly identified as not automatically generated (68\% of ``\textit{no}" to the question whether the output was produced by an artificial intelligence), followed by \gp~(54\%), and lastly by the Markov baseline (26\%), indicating, as expected, that the latter produces the least natural outputs. Figure~\ref{fig:class} reports the distribution over the various answers. Also in this case the distance between \gp~and gold is lower than \gp~and the baseline (double in percentage points), indicating that the production of \gp~is approaching natural language. 
It is also interesting to see that the highest percentage of ``\textit{can't tell}" is recorded for \gp, meaning that for this model it was harder than for baseline and gold to decide whether the text was automatic or not.

\begin{table*}[!t]
\centering
\begin{tabular}{l|ccc|ccc|ccc|ccc} \toprule
model & \multicolumn{3}{|c|}{5+5} &\multicolumn{3}{|c|}{5+10} &\multicolumn{3}{|c|}{10+5} & \multicolumn{3}{|c}{10+10}\\
\midrule
     & yes & no & ct & yes & no & ct & yes & no & ct & yes & no & ct  \\ \midrule

  Gold   & 26 & 66 & 8 & 27 & 68 & 5 & 32 & 63 & 5 & 28 & 71 & 1  \\ \midrule
  \gp    & 32 & 55 & 13 & 48 & 46 & 6 & 32 & 62 & 6 & 42 & 50 & 8 \\ \midrule
  Markov & 62 & 33 & 5 & 80 & 13 & 7 & 61 & 33 & 6 & 71 & 19 & 10 \\ 
\bottomrule
\end{tabular}
\caption{\label{tab:class}
Percentages of classification results according to the various stimulus material conditions. \\Is the text automatically generated? \textit{\{yes, no, can't tell (ct)\}}.}
\end{table*}

Let us look at results in more detail (Table~\ref{tab:class}), focusing again on length of input and continuation. Regarding continuation, we observe that *+5 conditions are  better than *+10 conditions for both automatic models, indicating that the least generated text, the more natural the fragment is perceived.

Regarding input length, we see that for \gp~a longer prompt yields better results (10+5 is better than 5+5, and 10+10 is better than 5+10). With 10-token prompts, \gp~generates text that is (i) assessed as natural as much as the original text when completed with 5 tokens (62\% \gp, 63\% original), and (ii) judged as natural 50\% of the times when completed with 10 tokens.
This seems to suggests that a longer input context is beneficial to \gp~when completion size is kept constant. However, we may wonder whether \gp~is evaluated as more natural because the generated text is actually better given the more context to start with, or simply because there is more gold text in the stimulus.
If it were just for the contribution of a longer gold portion in the stimulus, we should see a similar behaviour for the baseline. Instead, we see that prompt size doesn't matter for the baseline, at least for the 5 token completion case (33\% in both 5+5 and 10+5). In the 10-completions (5+10 and 10+10), the larger amount of gold data in the stimulus probably does alleviate a little the very low naturalness induced by the generated text. While we can tentatively postulate that \gp~generates better text when more input is provided, further investigation is required to provide more solid evidence.

%we have seen that best conditions for the baseline are with minimum automatic completion (5+5 and 10+5) independently of the prompt length. 

%that the Markov baseline is preferred when it adds the least amount of text to the original gold material (5+5 or 10+5 conditions). In the case of \gp, we observe instead a large influence of the initial (gold) material, with the best conditions being 10+5 and 10+10. 

%%% MARCO, ho rimosso questo paragafo perche' non ha piu' senso, con i nuovi risultati...
%Furthermore it seems that there is an `internal coherence' in the output produced by \gp~that can be derived by the fact that 5+5 is worse than 5+10, and that 10+10 is in the second position (while in the ranking scenario this was not evident). On the contrary Markov shows a consistent behavior with the ranking experiment, where the less the output added,  the better it is. 

%Markov: generation is bad. Indeed, 5+10 is worse than 5+5 and 10+10 is worse than 10+5. This indicates that given the same amount of gold, the more is added by the baseline, the easier it is to recognise the text as automatically generated.

%Geppetto: 5+5 is worse than 5+10

%GP effetto più forte è external coherence -> Prompt lungo = generazione migliore per geppetto in assoluto (come facciamo a sapere che il giudizio positivo non e' dato da extra gold?). 

%Secondo effetto di GP è internal coherence: dato un prompt corto la complition lunga è migliore 

\paragraph{Summary of Results.} Intersecting the observations from the two experimental setups provides us with a complete picture. 
In ranking (thus when the models are directly compared), both  \gp~and the baseline perform best in the 5+5 and 10+5 conditions, suggesting that automatic generation can easily be spotted when compared side by side with human text. In other words, the least generated material, the better. 

However, looking at classification, where each textual material is evaluated in isolation, we see that the two models behave in fact very differently. 
First, there is a much larger proportion of cases produced by \gp~that are deemed ``natural" (54\%) compared to Markov (26\%). Second, the margin of uncertainty when judging \gp~ is higher than for the baseline and for original text.  
Lastly, given the same completion size, \gp~performs better when its prompt is longer. Whether this is an effect of a larger proportion of gold data in the stimulus or it has to do with providing the model with a larger input context is left to future investigation.

% generated completion is longer  (5+10 and 10+10),\todo{double check} while Markov performs best when its completion is shorter, highlights the better \textit{coherence} of \gp. This is fostered both by a larger input context (external coherence), which explains the large gap between Markov and \gp~in the 10+* conditions, and by the longer portion of generated output (internal coherence), which explains why \gp~is much better than Markov at 5+10, compared to 5+5.\todo{check, not sure it makes sense}

% RANK PATTERN (ALL)
% gold	 gpt-2	markov	count	percentage
% 	1	  2	      3	    465	    38.750000
% 	2	  1	      3	    264	    22.000000
% 	1	  3	      2	    194	    16.166667
% 	3	  1	      2	    140	    11.666667
% 	2	  3	      1	    88	    7.333333
% 	3	  2	      1	    49	    4.083333

\section{Conclusion}

\gp~is the first GPT-2-based language model for Italian. Through both automatic and manual evaluation we assessed its quality on a variety of texts and in comparison to gold data as well as another statistical generation model.
Results show that \gp~is able
%, on one side, to model various linguistic styles (measured in terms of perplexity on corpora from different genres), and on the other 
to produce text which is much closer to human quality rather than to the text generated by the other generation model we have used. Linguistic analysis also highlights that \gp's production is quite similar to human production, though in a sort of bonsai version, since its sentences are on average shorter than the original texts, but with similar complexity.

The availability of \gp~opens up substantial possibilities. In the same way that GPT-2 is changing the approach to several NLP English tasks, we can expect \gp~to serve a similar purpose in Italian language processing.

\bibliography{acl2020}
\bibliographystyle{acl_natbib}

\end{document}